\documentclass{article} 
\usepackage{colm2024_conference}
\colmfinalcopy 
\usepackage{siunitx}  
\usepackage{float}
\usepackage{natbib}
\usepackage[skins,breakable]{tcolorbox}
\tcbuselibrary{skins}
\usepackage{geometry} 
\usepackage{tabularx}  
\usepackage{booktabs}  
\usepackage{makecell}  
\usepackage{ragged2e}
\usepackage[T1]{fontenc}
\usepackage{listings}
\usepackage{xcolor}
\usepackage{microtype}
\usepackage{hyperref}
\usepackage{url}
\usepackage{booktabs}
\definecolor{darkblue}{rgb}{0, 0, 0.5}
\hypersetup{colorlinks=true, citecolor=darkblue, linkcolor=darkblue, urlcolor=darkblue}

\definecolor{codegreen}{rgb}{0,0.6,0}
\definecolor{codegray}{rgb}{0.5,0.5,0.5}
\definecolor{codepurple}{rgb}{0.58,0,0.82}
\definecolor{backcolour}{rgb}{0.95,0.95,0.92}

\lstdefinestyle{mystyle}{
    backgroundcolor=\color{backcolour},   
    commentstyle=\color{codegreen},
    keywordstyle=\color{magenta},
    numberstyle=\tiny\color{codegray},
    stringstyle=\color{codepurple},
    basicstyle=\ttfamily\footnotesize,
    deletekeywords={next},
    breakatwhitespace=false,         
    breaklines=true,                 
    captionpos=b,                    
    keepspaces=true,                 
    numbers=left,                    
    numbersep=5pt,                  
    showspaces=false,                
    showstringspaces=false,
    showtabs=false,                  
    tabsize=2
}

\newtcolorbox{prompt}[1]{
    enhanced,
    drop shadow=black!5!white,
    left=4mm,
    right=4mm,
    top=2mm,
    bottom=2mm,
    boxsep=0mm,
    rounded corners,
    title=#1,
    fontupper=\footnotesize\linespread{0.9}\fontfamily{lmr}\selectfont,
    }

\title{LeetCodeDataset: A Temporal Dataset for Robust Evaluation and Efficient Training of Code LLMs}

\author{
Yunhui Xia\\
\texttt{newfacade@163.com}
\And
Wei Shen \thanks{Corresponding author} \\
\texttt{shenwei0917@126.com }
\And
Yan Wang\\
\texttt{wangyanps4@126.com}
\And
Jason Klein Liu \\
\texttt{jasonkleinlove@gmail.com}
\And
Huifeng Sun \\
\texttt{shelon\_2008@126.com}
\And
Siyue Wu \\
\texttt{wusy104@gmail.com}
\And
Jian Hu \\
\texttt{janhu9527@gmail.com}
\And
Xiaolong Xu \\
\texttt{xlxu@ieee.org}
}

\begin{document}

\maketitle

\begin{abstract}
We introduce LeetCodeDataset, a high-quality benchmark for evaluating and training code-generation models, addressing two key challenges in LLM research: the lack of reasoning-focused coding benchmarks and self-contained training testbeds. By curating LeetCode\footnote{\url{https://leetcode.com/}} Python problems with rich metadata, broad coverage, 100+ test cases per problem, and temporal splits (pre/post July 2024), our dataset enables contamination-free evaluation and efficient supervised fine-tuning (SFT). Experiments show reasoning models significantly outperform non-reasoning counterparts, while SFT with only 2.6K model-generated solutions achieves performance comparable to 110K-sample counterparts. The dataset and evaluation framework are available on Hugging Face\footnote{\url{https://huggingface.co/datasets/newfacade/LeetCodeDataset}} and Github\footnote{\url{https://github.com/newfacade/LeetCodeDataset}}.

\end{abstract}

\section{Introduction}

Code generation is critical in research and applications of large language models (LLMs). With the emergence of advanced reasoning models like OpenAI o1 \citep{o1-preview} and DeepSeek-R1 \citep{deepseekai2025deepseekr1incentivizingreasoningcapability}, two key challenges are highlighted.

The first challenge is the lack of coding benchmarks that accurately assess LLMs' reasoning abilities. LiveCodeBench \citep{jain2024livecodebenchholisticcontaminationfree}, a commonly used benchmark, addresses this by sourcing problems from platforms like LeetCode and AtCoder and using live updates to avoid data contamination. However, it has limitations: it covers a few problems per platform and lacks detailed tags for algorithms and data structures, making in-depth analysis difficult.

The second challenge is the absence of a self-contained testbed for training LLMs to master competition-level coding through methods such as supervised fine-tuning (SFT) \citep{zhou2024leveraging}, direct preference optimization (DPO) \citep{rafailov2023direct}, and reinforcement learning (RL), which are widely used for aligning model behavior with desired coding performance \citep{shen2024policy, shen2025exploring, hu2025reinforce++, liu2025adaptivestep}.
 While datasets such as APPS \citep{hendrycksapps2021}, CodeContests \citep{li2022competition}, and TACO \citep{li2023taco} provide competition problems split into training and test sets, they lack live updates and easy tools to support RL training workflows. Recently released Open-R1 CodeForces-CoTs \citep{penedo2025codeforces} dataset, generated by DeepSeek-R1, fails to filter solutions for correctness, limiting its reliability for rigorous skill evaluation.

To address these challenges, we introduce LeetCodeDataset, which fully leverages high-quality resources from LeetCode.  LeetCode is a popular online platform for coding practice and technical interview preparation. It offers over 3,000 algorithm and data structure problems at varying difficulty levels. The platform supports multiple languages (Python, Java, C++, etc.), providing real-time code testing with execution feedback. Developers use LeetCode to improve their problem-solving skills, prepare for tech company interviews, and join global programming competitions. We meticulously curated a LeetCode dataset covering over 90\% of Python problems on the platform. Each problem is annotated with rich metadata—including difficulty levels, release dates, and topic tags—and paired with 100+ test cases of varying complexity to minimize false positives. The dataset also includes an evaluation toolkit for fast and reliable assessment. To ensure temporal validity, we adopted a strict time-based split: problems released after July 1, 2024, form the test set for benchmarking, while those released earlier constitute the training set.

Using this dataset, we evaluated popular models—including proprietary and open-source models- and reasoning and non-reasoning architectures. Our evaluation shows that reasoning models outperform non-reasoning ones in competitive programming tasks, with Claude 3.7 Sonnet \citep{anthropic_2024_claude} performing best in its category. Additionally, we conducted supervised fine-tuning (SFT) training on the LeetCode training set. Despite using only 2.6K samples, the resulting model achieves performance comparable to counterparts trained on 110K code examples, demonstrating the exceptional training efficiency of the LeetCodeDataset.

\section{LeetCodeDataset}
\subsection{Data Collection}

As of the end of March 2025, the LeetCode platform hosted approximately 3,505 programming problems, among which 3,115 supported Python submissions. Our data collection process begins with this Python problem set, and we describe our process below.

\begin{figure}
  \centering
  \includegraphics[width=0.7\textwidth]{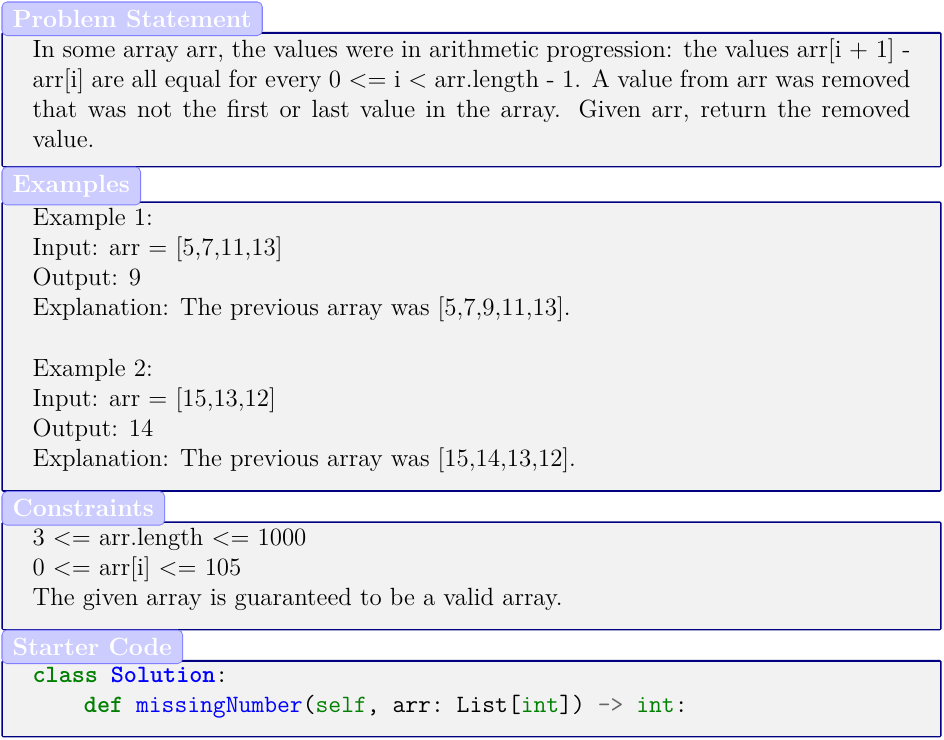} 
  \caption{An example of a LeetCode problem.} 
  \label{fig:example} 
\end{figure}

\textbf{Metadata Acquisition: } LeetCode provides a GraphQL API\footnote{\url{https://github.com/fspv/python-leetcode}} for accessing problem metadata and platform-hosted information. The following metadata fields were systematically collected for each problem: \textbf{\texttt{slug}} (URL identifier and primary key), 
\textbf{\texttt{question\_id}} (unique sequential number), 
\textbf{\texttt{difficulty}} (\textit{Easy}/\textit{Medium}/\textit{Hard}), 
\textbf{\texttt{problem\_description}} (full text, with examples and constraints, see \autoref{fig:example}), 
\textbf{\texttt{starter\_code}} (language template code), and 
\textbf{\texttt{topic\_tags}} (problem tags such as Array, Dynamic Programming).

\textbf{Canonical Solution Verification: } We retrieved reference solutions from various open-source GitHub repositories\footnote{\url{https://github.com/doocs/leetcode}}\footnote{\url{https://github.com/walkccc/LeetCode}}, and then verified the correctness of these solutions on the LeetCode platform, establishing ground truth solutions with a 100\% acceptance rate.

\textbf{Entry Point Identification: } The entry point refers to the function targeted for testing. In \autoref{fig:example}, this is \textit{missingNum}. Most starter codes contain a single function that is automatically identified as the entry point through text pattern matching. Specialized validation logic is necessary for problems requiring multiple functions (standard in design/simulation scenarios). However, such judgment codes are unavailable and challenging to develop. Therefore, our implementation focuses exclusively on single-function starter code scenarios.

\textbf{Input Generation: } To generate inputs for the entry point as part of test case development, we use one-shot prompting (\autoref{fig:prompt_input}) with the LLM. However, this method often produces overly simplistic inputs. To address this, we further prompt the LLM (\autoref{fig:prompt_complex_input}) to generate more complex inputs. By applying both approaches multiple times, we construct an average of over 100 inputs per problem, including many complex cases, significantly reducing the risk of false positives.

\textbf{Test Case Generation: } Now we have all the necessary information to generate test cases: specifically, we compute the Canonical Solution's entry point output using the previously generated inputs. To enable this, we developed a sandboxed execution environment for safe code evaluation, inserted the necessary imports before the canonical solution as part of the prompt, and handled special data structures such as binary trees (see \autoref{fig:import_binary_tree}) and linked lists (see \autoref{fig:import_linked_list}) separately.

After these steps, we successfully generated outputs for 2,869 problems, identifying the remaining cases as edge scenarios requiring additional investigation. Our pipeline ensures high dataset quality and comprehensive coverage, covering over 90\% of all Python problems available on the platform.

\textbf{LeetCodeDataset for SFT: } We designed LeetCodeDataset to serve dual purposes of model training and performance evaluation. The dataset employs a temporal split strategy: problems published after a predefined cutoff date (e.g., 2024-07-01) form our evaluation set, while earlier problems are allocated for supervised fine-tuning. The query of LeetCodeDataset is consistent with LiveCodeBench's construction \citep{jain2024livecodebenchholisticcontaminationfree}. For response generation, we intentionally avoided canonical solutions (often containing minimal comments or reasoning), which makes them suboptimal for instructional tuning. The detailed analysis can be found in \autoref{sec:efficient_training}. We employed Qwen2.5-Coder-32B-Instruct \citep{hui2024qwen2}, a highly sample-efficient and capable model, to implement a multi-stage generation process:

\begin{itemize}
    \item High-temperature sampling ($T=1.0$) produces diverse solution candidates.
    \item Automated test case verification filters functionally correct responses.
    \item For persistently failing problems, ground truth code snippets are integrated as contextual hints to improve the likelihood of correctness.
\end{itemize}

Finally, we developed the LeetCodeDataset, which features broad coverage, reliable benchmarking, evaluation/training splits based on release dates, and verified model-generated (query, response) pairs for SFT. The dataset can also support RL training by leveraging test cases as verifiers, making it a self-contained testbed for LLM development in code generation.

\subsection{Dataset Overview}

Now let's examine the constructed LeetCodeDataset. LeetCode problems can be categorized along multiple dimensions—we highlight three key ones below: difficulty, release date, and topic tags.

\textbf{Difficulty Levels:} As shown in \autoref{tab:difficulty_releaseyear}, LeetCode problems are categorized by difficulty into three levels:

\begin{itemize}
    \item \textbf{Easy}: Focuses on validating basic syntax and foundational data structure applications, typically solvable with straightforward logic.
    \item \textbf{Medium}: Requires familiarity with classical algorithms (e.g., dynamic programming, greedy) and the ability to design efficient strategies.
    \item \textbf{Hard}: Involves complex algorithmic combinations, mathematical insights, or specialized optimizations.
\end{itemize}

\begin{table}[!ht]
  \centering
  \begin{tabular}{l r r c l r r}
    \toprule
    \multicolumn{3}{c}{\textbf{Difficulty}} &  & \multicolumn{3}{c}{\textbf{Release Year}} \\
    \cmidrule{1-3} \cmidrule{5-7}
    \textbf{Type} & \textbf{Count} & \textbf{Proportion (\%)} & 
    & \textbf{Period} & \textbf{Count} & \textbf{Proportion (\%)} \\
    \midrule
    Easy   &  686  & 23.91 && Before 2020   & 1077 & 37.54 \\
    Medium & 1498  & 52.21 && 2020--2022    & 1009 & 35.17 \\
    Hard   &  686  & 23.88 && 2023--2025    &  783 & 27.29 \\
    \bottomrule
  \end{tabular}
  \caption{Distribution of difficulty and release year on the LeetCodeDataset.}
  \label{tab:difficulty_releaseyear}
\end{table}

\textbf{Release Date:} The release dates of LeetCode problems also offer valuable insights such as contamination-free evaluation of LLMs. Since LeetCode's weekly contest release dates and question IDs are publicly available, we use them as anchors to estimate each problem's release date. As shown in \autoref{tab:difficulty_releaseyear}, the yearly release distribution indicates approximately 350 new problems added annually in recent years. We argue that using problems from the past 6–12 months for benchmarking strikes an effective balance between bias and variance.

\begin{figure}[htbp]
  \centering
  \includegraphics[width=1.0\textwidth]{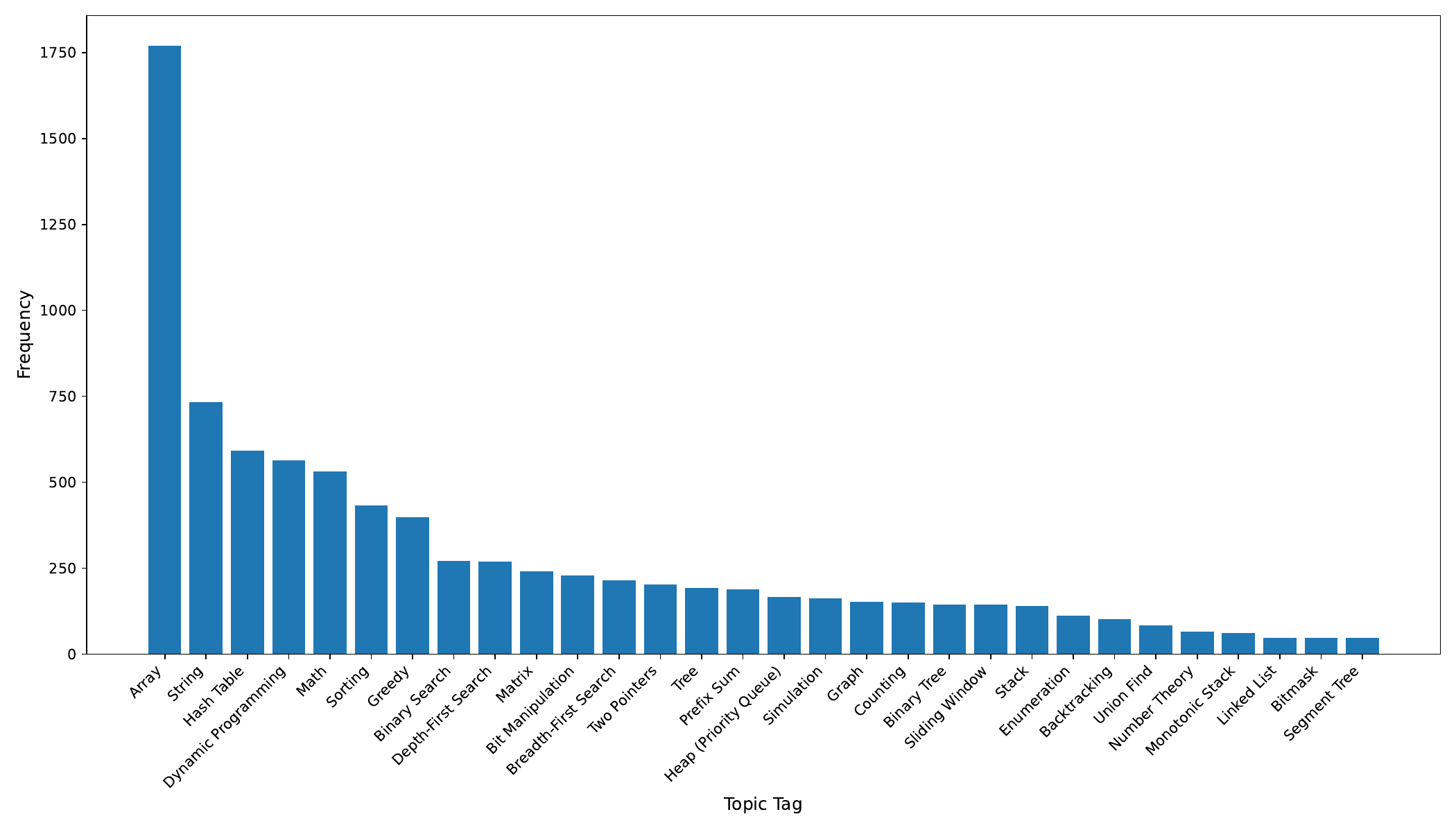} 
  \caption{Topic frequency distribution.} 
  \label{fig:topics} 
\end{figure}

\textbf{Topic Tags:} The LeetCode platform labels each problem with algorithm and data structure tags (e.g., Array, Binary Search), allowing multiple tags per problem. As shown in \autoref{fig:topics}, we examine how problems are distributed across these categories. This tagging system can help learners focus on specific skills. We believe this will provide insights to LLMs as well.

\section{Holistic Evaluation}

We evaluate six models on the LeetCodeDataset test set, comprising 256 programming problems that were newly released after July 1, 2024. The evaluated models include two proprietary systems, GPT-4o \citep{openai2024gpt4ocard} and Claude 3.7 Sonnet \citep{anthropic_2024_claude}; and four open-source models, DeepSeek-V3 \citep{deepseekai2025deepseekv3technicalreport}, DeepSeek-R1 \citep{deepseekai2025deepseekr1incentivizingreasoningcapability}, Qwen2.5-Max \citep{qwen25}, and QwQ-Plus \citep{qwq32b}. All experiments employ identical generation parameters with temperature=0.2 and top\_p=0.95 to ensure fair comparisons.

\begin{figure}[htbp]
  \centering
  \includegraphics[width=0.9\textwidth]{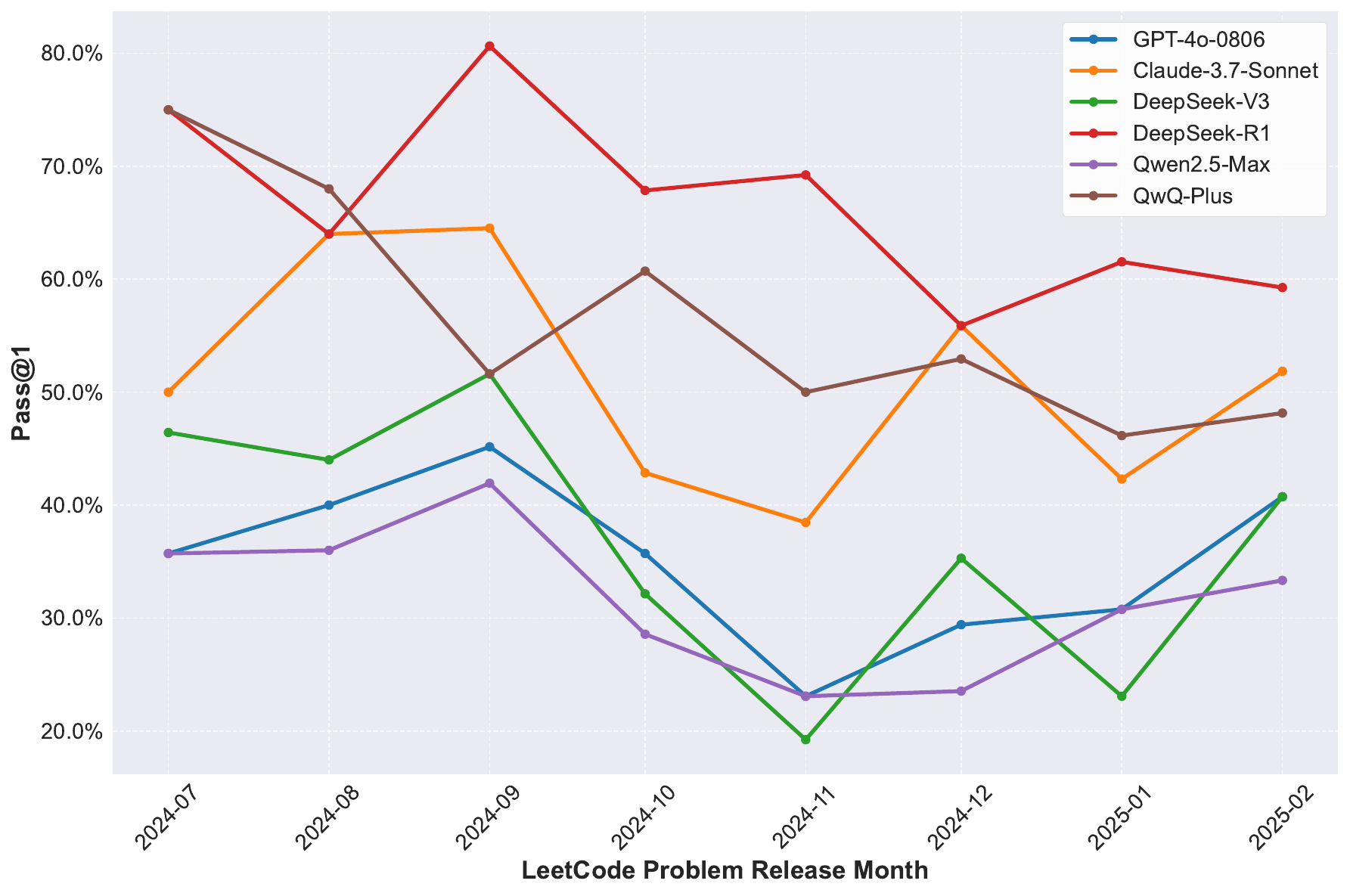} 
  \caption{Monthly pass rates of various models on the LeetCodeDataset.} 
  \label{fig:month_pass} 
\end{figure}

Following LiveCodeBench's temporal evaluation methodology, we analyze monthly accuracy change relative to problem release months as shown in \autoref{fig:month_pass},  and summarize model pass rates across difficulty levels in \autoref{tab:difficulty_pass}. This approach identifies potential data contamination by detecting declines in post-release accuracy, which would indicate overfitting to pre-release training data. Our findings reveal three key insights:

\begin{itemize}
\item \textbf{Superior Performance of Reasoning Models:} The evaluation highlights DeepSeek-R1 (pass@1 rate = 65.23\%) and QwQ-Plus (pass@1 rate = 56.25\%) as top performers, demonstrating the substantial advantage of long-CoT reasoning models in solving complex competition-level coding problems.
\item \textbf{Baseline Comparison:} Claude-3.7-Sonnet, operating without extended thinking, achieves superior performance within its model category. The two models, GPT-4o and DeepSeek-V3, achieved the same overall score. GPT-4o performs slightly better on easy problems, while DeepSeek-V3 performs slightly better on hard problems.
\item \textbf{Contamination Analysis:} The minimal temporal overlap between GPT-4o-0806's release date (August 2024) and our test problem release window (post-July 2024) strongly suggests authentic model capability measurements. We see similar curves among GPT-4o-0806, DeepSeek-V3, and Qwen2.5-Max; we believe the monthly accuracy fluctuations are mainly due to changes in problem
difficulty.
\end{itemize}

\begin{table}[!ht]
\centering
\begin{tabular}{@{} l *{4}{S[table-format=2.2]} @{}}
\toprule
\textbf{Model}             & \textbf{Easy (\%)} & \textbf{Medium (\%)} & \textbf{Hard (\%)} & \textbf{Overall (\%)} \\
\midrule
GPT-4o-0806       & 81.48 & 32.76 & 10.47 & 35.55 \\
Claude-3.7-Sonnet & 87.04 & 54.31 & 23.26 & 50.78 \\
DeepSeek-V3       & 77.78 & 31.90 & 13.95 & 35.55 \\
DeepSeek-R1       & 94.44 & 68.97 & 41.86 & 65.23 \\
Qwen2.5-Max       & 74.07 & 25.00 & 10.47 & 30.47 \\
QwQ-Plus          & 92.59 & 62.93 & 24.42 & 56.25 \\
\bottomrule
\end{tabular}
\caption{Model pass rates by difficulty level on the LeetCodeDataset.}
\label{tab:difficulty_pass}
\end{table}

We also analyze model pass rates across different topic tags, as depicted in \autoref{tab:topic_tag_eval}. By comparing these results, we identify each model's strengths and weaknesses, which provides insights for future improvements. Our key findings include:

\begin{itemize}
    \item The reasoning model DeepSeek-R1 shows strong performance across all topic tags, with pass rates mostly ranging from 60\% to 70\% and minimal variation. In contrast, non-reasoning models like GPT-4o exhibit significant fluctuations, such as dropping to 7.7\% in Binary Search tasks but reaching 63.2\% in Simulation tasks.
    \item We observe significant performance differences between reasoning and non-reasoning models in Dynamic Programming, Binary Search, and Tree-related tasks. This pattern demonstrates the need for additional reasoning capabilities in these domains.
\end{itemize}

\begin{table}[htbp]
\centering
\footnotesize 
\setlength{\tabcolsep}{4pt} 
\begin{tabularx}{\textwidth}{@{}l *{6}{>{\centering\arraybackslash}X}@{}}
\toprule
                      & \makecell{GPT-4o} 
                      & \makecell{DeepSeek\\-V3} 
                      & \makecell{Qwen2.5\\-Max} 
                      & \makecell{Claude-3.7\\-Sonnet} 
                      & \makecell{DeepSeek\\-R1} 
                      & \makecell{QwQ\\-Plus} \\
\midrule
Array                 & 32.1   & 34.5        & 28.0        & 51.2              & 67.9        & 55.4     \\
String                & 37.3   & 38.8        & 35.8        & 49.3              & 68.7        & 50.7     \\
Dynamic Programming   & 10.5   & 15.8        & 8.8         & 31.6              & 70.2        & 40.4     \\
Hash Table            & 39.5   & 37.5        & 35.7        & 50.0              & 66.1        & 50.0     \\
Math                  & 38.2   & 40.0        & 32.7        & 56.4              & 69.1        & 58.2     \\
Greedy                & 12.5   & 15.6        & 12.5        & 21.9              & 62.5        & 28.1     \\
Sorting               & 20.0   & 20.0        & 6.7         & 36.7              & 66.7        & 53.3     \\
Prefix Sum            & 17.9   & 14.3        & 14.3        & 35.7              & 71.4        & 35.7     \\
Binary Search         & 7.7    & 23.1        & 11.5        & 30.8              & 73.1        & 30.8     \\
Sliding Window        & 52.2   & 47.8        & 43.5        & 69.6              & 56.5        & 52.2     \\
Enumeration           & 27.3   & 31.8        & 9.1         & 45.5              & 63.6        & 50.0     \\
Matrix                & 19.0   & 33.3        & 19.0        & 52.4              & 76.2        & 61.9     \\
Simulation            & 63.2   & 57.9        & 42.1        & 63.2              & 63.2        & 84.2     \\
Depth-First Search    & 31.6   & 21.1        & 26.3        & 31.6              & 57.9        & 57.9     \\
Bit Manipulation      & 33.3   & 44.4        & 27.8        & 50.0              & 50.0        & 66.7     \\
Combinatorics         & 12.5   & 18.8        & 12.5        & 37.5              & 93.8        & 25.0     \\
Counting              & 20.0   & 26.7        & 26.7        & 46.7              & 53.3        & 46.7     \\
Graph                 & 40.0   & 33.3        & 46.7        & 53.3              & 66.7        & 66.7     \\
Heap (Priority Queue) & 40.0   & 53.3        & 33.3        & 66.7              & 66.7        & 66.7     \\
Number Theory         & 38.5   & 30.8        & 30.8        & 38.5              & 69.2        & 53.8     \\
Breadth-First Search  & 41.7   & 33.3        & 50.0        & 58.3              & 58.3        & 75.0     \\
Tree                  & 27.3   & 18.2        & 9.1         & 9.1               & 72.7        & 54.5     \\
Two Pointers          & 20.0   & 30.0        & 30.0        & 40.0              & 80.0        & 40.0     \\
Segment Tree          & 30.0   & 30.0        & 30.0        & 70.0              & 80.0        & 30.0     \\
All                   & 35.5   & 35.5        & 30.5        & 50.8              & 65.2        & 56.2     \\
\bottomrule
\end{tabularx}
\caption{Pass rates of models across topic tags.}
\label{tab:topic_tag_eval}
\end{table}

\section{Efficient Training}
\label{sec:efficient_training}

\subsection{Experiment Setup}

We conducted SFT using Qwen2.5-Coder-7B \citep{hui2024qwen2} as our base model. The model was trained for 3 epochs with an initial learning rate of 1e-5, employing a warmup ratio of 0.1 and cosine learning rate scheduling. All experiments utilized consistent hyperparameters, including a batch size of 32 across different datasets.

\subsection{Results}

To evaluate the training efficiency of LeetCodeDataset, we conducted comparative experiments with five widely-used coding datasets (\citealp{wei2024magicoderempoweringcodegeneration}; \citealp{luo2023wizardcoderempoweringcodelarge}; 
 \citealp{penedo2025codeforces}; \citealp{openthoughts}) ranging from 9.5K to 111.1K samples - all substantially larger than our LeetCodeDataset training set. Under identical experimental configurations above, we trained models on each dataset and evaluated them across four benchmarks: HumanEval \citep{chen2021evaluatinglargelanguagemodels}, MBPP \citep{austin2021programsynthesislargelanguage}, LiveCodeBench \citep{jain2024livecodebenchholisticcontaminationfree}, alongside our newly developed LeetCodeDataset evaluation set. As demonstrated in \autoref{tab:sft_train}, we summarize our key findings:

\begin{itemize}
    \item \textbf{Superior Model-Generated Training Data.} The SFT-trained model using model-generated responses from the pre-2024-07 LeetCodeDataset significantly outperformed the version trained on human-written responses (79.9\% vs. 55.5\% on HumanEval; 77.5\% vs. 53.4\% on MBPP), despite both response types being verified as correct. The result highlights the quality advantage of model-generated training data for code generation tasks.
    \item \textbf{High Data Efficiency.} Training with only 2.6K model-generated LeetCode samples achieved superior performance on HumanEval (79.9\%) and MBPP (77.5\%), surpassing models trained on much larger datasets (9.5K–111.1K rows). The finding demonstrates exceptional data efficiency for domain-specific code generation.
    \item \textbf{Limitations on Hard Benchmarks.} Despite being in-distribution for LeetCodeDataset (post-2024-07), the 2.6K-trained model underperformed on hard benchmarks. It suggests that small-scale SFT primarily develops basic programming skills.
\end{itemize}

\begin{table}[htbp]
\centering
\setlength{\tabcolsep}{4pt} 
\begin{tabularx}{0.95\textwidth}{ 
  >{\raggedright\arraybackslash}p{3.5cm} 
  *{5}{>{\centering\arraybackslash}X} 
}
\toprule
\multicolumn{1}{l}{\textbf{Training Data}} & 
\multicolumn{1}{c}{\textbf{\makecell{Rows}}} & 
\multicolumn{1}{c}{\textbf{\makecell{Human\\Eval}}} & 
\multicolumn{1}{c}{\textbf{\makecell{MBPP}}} & 
\multicolumn{1}{c}{\textbf{\makecell{LiveCode\\Bench\\24-08$\sim$25-02}}} & 
\multicolumn{1}{c}{\textbf{\makecell{LeetCode\\Dataset\\24-07$\sim$25-03}}} \\
\midrule
\makecell[l]{Magicoder\\Evol-Instruct-110K} & 111.1K & 77.4 & 74.1 & 15.1 & 13.7 \\
\addlinespace[2pt]
\makecell[l]{Magicoder\\OSS-Instruct-75K}   & 75.1K  & 73.8 & 76.5 & 15.1 & 12.9 \\
\addlinespace[2pt]
\makecell[l]{Open-R1\\CodeForces-CoT}       & 9.5K   & 79.9 & 74.1 & 15.8 & 13.3 \\
\addlinespace[2pt]
\makecell[l]{OpenThoughts\\114k}            & 19.9K  & 77.4 & 75.7 & 16.9 & 16.4 \\
\addlinespace[2pt]
\makecell[l]{LeetCodeDataset\\Pre 2024-07 human} & 2.6K & 55.5 & 53.4 & 14.0 & 10.9 \\
\addlinespace[2pt]
\makecell[l]{LeetCodeDataset\\Pre 2024-07 model} & 2.6K & 79.9 & 77.5 & 15.4 & 12.5 \\
\bottomrule
\end{tabularx}
\caption{Model SFT-training results.}
\label{tab:sft_train}
\end{table}

\section{Related Work}

\textbf{Code Generation Benchmarks.} Numerous benchmarks have been developed to evaluate the code generation capabilities of LLMs. For foundational Python programming, widely used benchmarks include HumanEval \citep{chen2021evaluatinglargelanguagemodels} and MBPP \citep{austin2021programsynthesislargelanguage}. EvalPlus \citep{liu2023codegeneratedchatgptreally} offers a more rigorous variant. Multiple-E \citep{cassano2022multiplescalableextensibleapproach} further extends these two popular benchmarks by translating them into 18 other programming languages. As LLM capabilities advance, many of these benchmarks are becoming too easy to assess modern models adequately. A few specialized benchmarks focus on competitive programming challenges. APPS \citep{hendrycksapps2021}, CodeContests \citep{li2022competition}, and TACO \citep{li2023taco} source problems from platforms like Codeforces and AtCoder. LiveCodeBench \citep{jain2024livecodebenchholisticcontaminationfree} provides holistic and contamination-free evaluations by dynamically updating coding challenges from platforms like LeetCode and AtCoder. CODEELO \citep{quan2025codeelobenchmarkingcompetitionlevelcode} tries to align with the CodeForces platform by submitting directly to the platform and developing an Elo rating calculation system.

\textbf{Fine-tuning Dataset of Code.} Synthetic data is one primary source of LLM SFT data. CodeAlpaca \citep{codealpaca} employs few-shot prompting and teacher models to synthesize data for code-specific fine-tuning. Magicoder \citep{wei2024magicoderempoweringcodegeneration} leverages open-source code snippets to generate high-quality instructional data for coding tasks. In competitive programming benchmarks like APPS and CodeTest, training splits are provided for SFT, utilizing competition-level problems to enhance model problem-solving capabilities. For advanced reasoning, pen-R1 CodeForces-CoTs \citep{penedo2025codeforces} includes 10K CodeForces problems with up to five reasoning traces generated by DeepSeek R1. In contrast, OpenThoughts \citep{openthoughts} is a synthetic dataset with 114K high-quality examples spanning math, science, code, and puzzles.

\section{Limitations}

While our LeetCode dataset effectively benchmarks and fine-tunes code models, it has three key limitations:

\textbf{False Positive Risks:} Though we designed diverse inputs and test cases to reduce incorrect solutions passing, our dataset lacks extremely complex input patterns and suffers from an imbalanced test case distribution. These limitations present residual risks of false positives (e.g., solutions passing tests despite logic errors).

\textbf{Complexity Analysis Gap:} Determining time/space complexity for problems requires LeetCode-style test cases tailored to each algorithm’s behavior. The limitation exceeds our current scope as it demands manual problem-specific validation.

\textbf{Coverage Gaps:} We haven’t included certain problem types, particularly problems with multiple solution entry points.

\section{Conclusion}

We present LeetCodeDataset, a rigorously curated resource that addresses key challenges in code-generation research for large language models. By aggregating 2,869 Python LeetCode problems—each annotated with rich metadata (difficulty, tags, release dates) and augmented with 100+ diverse test cases—our dataset enables reliable, contamination-free model evaluation and highly efficient training. Its temporal split (with post-July 2024 problems as the test set) ensures clean benchmarking and supports longitudinal studies.
This dataset comprehensively covers algorithms and data structures, facilitating robust overall evaluation and fine-grained skill analysis. With an integrated evaluation toolkit, LeetCodeDataset streamlines assessment and comparison across models.
Notably, we show that models trained on just 2.6K curated samples from LeetCodeDataset can match the performance of those trained on 110K examples from previous benchmarks, demonstrating strong data efficiency. We expect LeetCodeDataset to become a foundational resource for developing, training, and evaluating advanced code-generation models.


\bibliography{colm2024_conference}
\bibliographystyle{colm2024_conference}

\appendix
\section{Appendix}
\subsection{Prompts}

During input generation for entry points, we sample three tree, three linked list, and four other problem types, extracting input specifications from their descriptions to define entry points. These 10 selected problems serve as one-shot examples in the Input-Generation-Prompt, with domain-specific constraints: tree problems use only tree examples; linked list problems draw from linked list cases; others follow the same principle, ensuring generated inputs align with each problem type's structural requirements.

\begin{figure}[H]
\centering
\begin{prompt}{Input-Generation-Prompt}
\small
You are an expert Python programmer. You will be given a question (including a problem specification and starter code). Your task is to generate inputs that are consistent with the problem specification and starter code. An example will be provided for illustration.
\\
\\
**** Example ****
\\
\\
\#\#\#\# Question:

{\color{blue}\{example problem description and starter code\}}
\\
\\
\#\#\#\# Some valid inputs of the starter code (json format):

\textasciigrave\textasciigrave\textasciigrave json

{\color{blue}\{example problem inputs\}}

\textasciigrave\textasciigrave\textasciigrave
\\
\\
**** Now Your Task ****
\\
\\
\#\#\#\# Question:

{\color{blue}\{problem description and starter code\}}
\\
\\
\#\#\#\# Some valid inputs of the starter code (json format):
\end{prompt}
\caption{Prompt structure for input generation.}
\label{fig:prompt_input}
\end{figure}

\begin{figure}[H]
\centering
\begin{prompt}{Complex-Input-Generation-Prompt}
You are an expert Python programmer. You will be given a question (including a problem specification and starter code) along with a few sample inputs. Your task is to generate additional inputs that are consistent with the question and the provided sample inputs.
\\
\\
\#\#\#\# Question:

{\color{blue}\{problem description and starter code\}}
\\
\\
\#\#\#\# Sample inputs (using json format):

\textasciigrave\textasciigrave\textasciigrave json

{\color{blue}\{sample inputs\}}

\textasciigrave\textasciigrave\textasciigrave
\\
\\
\#\#\#\# Generate some additional inputs that are more complex than the sample inputs (using json format):
\end{prompt}
\caption{Prompt structure for complex input generation.}
\label{fig:prompt_complex_input}
\end{figure}

\subsubsection{Handle Data Structures}

To ensure robust evaluation, we prepend essential imports (e.g., \verb|from typing import List|) to all code completions. Special handling is required for binary tree and linked list data structures, which involve additional utility functions for serialization/deserialization. Below are the supplementary imports and helper functions used to manage these structures:

\begin{figure}[H]
\begin{lstlisting}[language=Python]
from typing import Optional
from collections import deque


class ListNode:
    def __init__(self, val=0, next=None):
        self.val = val
        self.next = next


def list_node(values: list) -> Optional[ListNode]:
    if not values:
        return None
    head = ListNode(values[0])
    p = head
    for val in values[1:]:
        node = ListNode(val)
        p.next = node
        p = node
    return head


def linked_list_to_list(head: Optional[ListNode]) -> list:
    result = []
    current = head
    while current:
        result.append(current.val)
        current = current.next
    return result


def is_same_list(p1: Optional[ListNode], p2: Optional[ListNode]) -> bool:
    if p1 is None and p2 is None:
        return True
    if not p1 or not p2:
        return False
    return p1.val == p2.val and is_same_list(p1.next, p2.next)
\end{lstlisting}
\caption{Additional imports related to linked list2.}
\label{fig:import_linked_list}
\end{figure}

\begin{figure}[H]
\begin{lstlisting}[language=Python]
from typing import Optional
from collections import deque


class TreeNode:
    def __init__(self, val=0, left=None, right=None):
        self.val = val
        self.left = left
        self.right = right


def tree_node(values: list) -> Optional[TreeNode]:
    if not values:
        return None
    root = TreeNode(values[0])
    i = 1
    queue = deque()
    queue.append(root)
    while queue:
        node = queue.popleft()
        if i < len(values) and values[i] is not None:
            node.left = TreeNode(values[i])
            queue.append(node.left)
        i += 1
        if i < len(values) and values[i] is not None:
            node.right = TreeNode(values[i])
            queue.append(node.right)
        i += 1
    return root


def tree_node_to_list(root: Optional[TreeNode]) -> list:
    if not root:
        return []

    result = []
    queue = deque()
    queue.append(root)

    while queue:
        node = queue.popleft()
        if node:
            result.append(node.val)
            queue.append(node.left)
            queue.append(node.right)
        else:
            result.append(None)

    while result and result[-1] is None:
        result.pop()

    return result


def is_same_tree(p: Optional[TreeNode], q: Optional[TreeNode]) -> bool:
    if not p and not q:
        return True
    elif not p or not q:
        return False
    elif p.val != q.val:
        return False
    else:
        return is_same_tree(p.left, q.left) and is_same_tree(p.right, q.right)
\end{lstlisting}
\caption{Additional imports related to binary tree.}
\label{fig:import_binary_tree}
\end{figure}

\end{document}